\documentclass[12pt]{article}

\usepackage{rotating}

\usepackage{tikz}
\usepackage[utf8]{inputenc}
\usepackage[T1]{fontenc}
\usepackage{amsmath}
\usetikzlibrary{fit,positioning}

\usepackage{subfig}
\usepackage[ruled,vlined,linesnumbered]{algorithm2e}
\usepackage{multirow}
\usepackage{ulem}
\usepackage{soul}
\usepackage{booktabs,tabu}
\usepackage{bm}
\usepackage{array}
\usepackage{enumerate}
\usepackage{url}
\usepackage{amsfonts}
\newcolumntype{C}{@{\extracolsep{1cm}}c@{\extracolsep{2pt}}}%

\allowdisplaybreaks
\begin{document}

\title{Pre-Trained Language Transformers are Universal Image Classifiers}

\author{Rahul Goel$^\star$, Modar Sulaiman$^\star$, Kimia Noorbakhsh$^\dagger$, Mahdi Sharifi$^\ddagger$\\
Rajesh Sharma$^\star$, Pooyan Jamshidi$^\ddagger$, Kallol Roy$^\star$\\
$^\star$ University of Tartu, Estonia\\
$^\dagger$ Sharif University of Technology, Iran\\
$^\ddagger$ University of South Carolina, U.S.}

\date{}
\maketitle

\begin{abstract}
Facial images disclose many hidden personal traits such as age, gender, race, health, emotion, and psychology. Understanding these traits will help to classify the people in different attributes. In this paper, we have presented a novel method for classifying images using a pretrained transformer model. We apply the pretrained transformer for the binary classification of facial images in criminal and non-criminal classes. The pretrained transformer of GPT-2 is trained to generate text and then fine-tuned to classify facial images. During the finetuning process with images, most of the layers of GT-2 are frozen during backpropagation and the model is frozen pretrained transformer (FPT). The FPT acts as a universal image classifier, and this paper shows the application of FPT on facial images. We also use our FPT on encrypted images for classification. Our FPT shows high accuracy on both raw facial images and encrypted images. We hypothesize the meta-learning capacity FPT gained because of its large size and trained on a large size with theory and experiments. The GPT-2 trained to generate a single word token at a time, through the autoregressive process, forced to heavy-tail distribution. Then the FPT uses the heavy-tail property as its meta-learning capacity for classifying images. Our work shows one way to avoid bias during the machine classification of images. The FPT encodes worldly knowledge because of the pretraining of one text, which it uses during the classification. The statistical error of classification is reduced because of the added context gained from the text. Our paper shows the ethical dimension of using encrypted data for classification. Criminal images are sensitive to share across the boundary but encrypted largely evades ethical concern. FPT showing good classification accuracy on encrypted images shows promise for further research on privacy-preserving machine learning. We have collected facial images of criminal and non-criminal classes from multiple web sources for a balanced training data set. For experimentation, we used in total 20k images of criminals and non-criminals (10k for each class) collected from eleven different data sources. Our findings show that our proposed model can classify criminal and non-criminal images with 0.99 AUROC and 0.98 average Precision. For better generalization of our model, we also encrypted images using the Chaos-Based Image-Encryption algorithm. In our experiments, the proposed method can classify encrypted images with 0.98 AUROC and 0.97 average Precision.
\end{abstract}

\section{Introduction}\label{sec:Introduction}

Vision is a vital modality for humans to understand the complex world around them \cite{goldberg2009new}. In modern-day lives, the police force, security agencies, defense agencies use images of a person to understand their criminal tendency \cite{hashemi2020retracted}. As goes with the folklore, "a picture speaks thousands of words", we can determine people's feelings, emotions, aggressiveness just by looking at their face with fairly good accuracy. The sensitivity of detecting a person's criminal tendency from their face in railway stations, airports, mass gatherings, immigration demands very high accuracy. Security persons monitor sensitive places by looking at the crowd through computer monitors or in persons. Any chance of detection error may have huge bearings on people's lives. On top of it, defense forces nowadays rely heavily on automated machine learning systems to detect criminal tendencies from facial images. This comes from the practical necessity of detecting images at scale in the crowded areas. Convolution neural networks (CNN) are employed as classifiers in the automated classification of images to binary classes, e.g., (Class 0 = criminal, Class 1 = non-criminal). Though CNN acts as a good classifier, they use only the features of images to classify \cite{hatami2018classification}. For example, if CNN sees many squarish-faced, thick nose images as criminal people in the training set, the CNN will be biased towards labeling similar features in the test set as criminal. This inductive bias of CNN comes both from the combination of CNN's architectural constraints and the imbalanced distributions of the training data. The architectural constraints of CNN force to combine the low-level features of images and pass them to higher layers. The CNN generally used for the classification are powerful pattern matchers and has the ability to contort themselves to fit almost any unbalanced facial images dataset. CNN model understands the world by gluing together thousands (even millions) of linear and non-linear functions, and adapting each of these functions using a training example. It thus builds a high dimensional manifold that fits the training image data set, and generalizes by inductively pattern-matching onto what it has seen before in the training dataset. CNN (and a wide variety of machine learning models) thus generalize their prediction by induction and not by deduction. This explains why big deep vision models require large amounts of data to learn to classify. Big vision models are good inductive interpolators but not so good extrapolators. Thus, deep vision models of CNN (Capsule Networks) can be risky to classify the images of people, and these models' understanding of the images is superficial. Moreover, the limited availability of criminal images compared to non-criminal images add further problem. To better understand the facial images, we need to give them additional context in the form of architectural priors and statistical priors. Building image priors for CNN is an arduous task, and the deep learning community relies on data at scale for better generalization.

In this paper, we circumvent the above problems of deep vision models for binary classification of images using a novel method of transfer learning. We use the language model of the Frozen Pre-Trained Transformer (FPT) model for the classification task. Our FPT is pre-trained on a vast amount of text, and then the learned FPT is fine-tuned on facial images for classification. FPT is pre-trained on publicly available data sets, e.g., Wikipedia, Google books, etc., to generate the next word sequence through the autoregressive method. The autoregressive pre-training method gives the heavy-tailed distribution of the output generated words. The heavy-tailed behavior gained by FPT during word generation act as a prior for the fine-tuning process of the image classification. Because of the pre-training, our FPT bypasses the limitations of the inductive assumption that the unseen test data will resemble the training data. Though this is true in some cases, it does not always hold. Facial images with similar features in the training and test sets may tell a completely different story. The model thus may wrongly interpret the unseen test image. We have used the autoregressive pre-training method to overcome the inductive assumption, and our FPT model the world in a non-linear fashion. The FPT may predict similar images in the training and test data differently. The long-tail distribution helps the model to catch the criminal cases that are less common (e.g., old white women). These training examples of these outliers and the hardest to come across for the FPT to train. The heavy-tail also assists FPT to circumvent different biases (racial, ethnic, facial features) during classification by non-linear modeling of training and test data. To the best of our knowledge, this paper is the first attempt to use the language model to detect the criminal tendency from a facial image. 

The main contributions of our paper are as follows: 

\begin{itemize}
  \item We use a pre-trained language model to classify facial images for criminal detection.
  \item We have theoretically proved and experimentally validated that Frozen Pre-Trained Transformer (FPT) is an \textit{Universal Image Classifier}.
  \item We have shown experimentally that FPT can classify encrypted images. 
  \item We built a balanced facial images data set by collecting the images from 11 different sources through web scraping. 
\end{itemize}

\section{Related work}
Research has shown that people judge a person by their facial features \cite{todorov2015social,zhao2003face}. People infer whether a person is likely to be trustworthy, competent, or dominant by looking at photos or computer-generated images \cite{buckingham2006visual}. Psychological research shows we judge a person when we see him/her first time, within a couple of seconds. The long history of evolution has trained humans (and other primates, animals) to quickly infer its danger, risk, etc., by looking at his/her group members' facial features. Present vision systems have developed amazingly to detect a person's trustworthiness, honesty, etc., simply by looking at the face \cite{goel2021facial,bertini2015profile}. The police and defense agencies have largely relied on the ingenuity of human vision systems to detect criminals—robbers and people who choose to live shady life, unfortunately. Moreover, the criminals are also getting smarter every day and become quite successful in evading the tactics of the police\cite{buckingham2006visual}. Thus, there always lies a finite probability of a  \textbf{Type I} error (false positive conclusion) and \textbf{Type II error} (false negative conclusion)\cite{haselhuhn2012bad}. Though the initial impression comes from facial images, they are inherently influenced by bias (racial, ethnic, color, personal). Multiple sources of information need to be fused to overcome the embedded bias\cite{todorov2015social,stoker2016facial}. This complexity brings many challenges to empirical studies of facial images. The criminal manifold and non-criminal manifold are mixed, and it is a nontrivial task to separate them. 

The criminal tendency has been thoroughly investigated in work by Lombroso. His works on \textit{"born criminal"} influenced European and American thinking about the causes of criminal behavior. It also laid the foundations of theories of crime, explained by the facial structures and genetic explanations\cite{lombroso2006criminal}. Lombroso's work put the research perspective of facial structure and emotions entangling to the forefront. In \cite{wu2016automated}, authors study the possibility of teaching machines to pass the Turing test on the task of duplicating humans in their first impressions with facial images, personality traits etc. The thesis of their study is to face the perception of criminality and to validate the hypothesis on the correlations between the innate traits of a person and his/her facial features. A flurry of research activity of using deep neural networks for computer vision challenges started from the paper of Alex Krizhevsky et al. in the year of 2012\cite{krizhevsky2012imagenet}. Deep neural networks, mainly convolutional networks, capsule networks, and recently vision transformers, are ubiquitous use in computer vision research\cite{taneja2019modeling}. Neural networks extract the features from facial images data only\cite{ouyang2015deepid,sun2015deepid3}. These features then passed through non-linear transform functions to classify sexual orientation from facial images. The features to identify sexual orientation are learned from the data exclusively through backpropagation and have shown the accuracy of  81\% for men and 71\% for women\cite{wang2018deep}. The accuracy of their deep neural systems surpasses human judges, showing the accuracy of 61\% for men and 54\% for women. Xin Geng et al. estimate the age through facial images using algorithms of IIS-LLD and CPNN\cite{geng2013facial}. The authors exploited the changes of facial features as a slowly varying function of time to overcome the scarcity of training data. In \cite{geng2013facial}, the authors estimate the age through facial images. Along similar lines, Andrew G Reece et al. used Instagram face images as markers to detect depression and psychiatric disorders\cite{reece2017instagram}. Using color analysis metadata components, they compute multiple statistical features from the Instagram photos. Their proposed models also outperformed humans' success rate for identifying depression. The authors of \cite{chackravarthy2018intelligent} presented a method for detecting criminal activity in a video stream by recording the person's aberrant actions in subsequent video frames. They have employed a hybrid deep learning algorithm to analyze video stream data for surveillance in urban residential areas. Umadevi V Navalgund et al presented a fascinating work of capturing crime intention in public places of ATM, Bank etc by detecting the weapons in hand\cite{navalgund2018crime}. This is in the direction of pre-crime technology. Where the machine learning model tries to predict the crime before it happens. The authors have used pre-trained deep models of GoogleNet, VGGNet-19 for the predictive behavior. In closely related works, pre-trained deep learning models such as VGG-19 \cite{mateen2019fundus}, and GoogleNet \cite{zhong2015high} have been used to identify a knife or pistol in a person's hand and aim it towards another person. The authors of \cite{rajapakshe2019using} advocated real-time criminal detection utilizing ML and Deep Learning for crime prevention. The use of pre-trained deep learning models e.g. VGG-16, VGG-19, GoogleNet, and Inception V3 is to train on a wider variety of data set\cite{verma2020convolutional,mateen2019fundus,zhong2015high}. The pre-trained models thus encode worldly knowledge through pre-training, which is used to detect criminal tendencies/intentions from facial images, crowd images, etc. As the deep models build their features from the data (without any hand coding), its unreasonable effectiveness (accuracy in predicting criminal tendencies) comes from the volume and veracity of data.

\section{Theory}
The FPT is a transformer-decoder block trained on massive web text crawled from the internet. The FPT is trained to generate a one-word token at a time. Our FPT architecture, unlike a normal transformer, has no encoder. For example, at the start, the output word token is $``the"$ that generates the next word token ``cat''. The new sequence is now $[``the", ``cat"]$ fed in FPT to generate the next word token $``mat"$. This is an autoregressive method by which FPT is pretrained on the web text to generate a single word token every time. We use this autoregressive pre-trained model to generate a word token as a stochastic recurrence equation. The word token $X_{t}$ generated at time $t$ depends on $X_{t-1}$ as\cite{buraczewski2018stochastic, MR0440724}:

\begin{equation} \label{eq1}
X_{t}  = A_{t}X_{t-1} + B_{t} \\
\end{equation}
where $A_{t}$ and $B_{t}$ matrices comes from the FPT architecture. The solution for Equation~\ref{eq1} is then given by the  recursive method as:
\begin{equation}\label{eq2}
X_{t} = B_{t} + A_{t}B_{t-1} + A_{t}A_{t-1}B_{t-2} + \cdots + A_{t}A_{t-1} \cdots A_{2}B_{1} + A_{t}A_{t-1}\cdots A_{1}X_{0}
\end{equation}
Thus word token generated at time $X_{t}$ depends on the initial word token $X_{0}$ and $\mathbb{P}(X_{t})$ follows:

\begin{equation}\label{eq3}
\Sigma_{k=1}^{t}A_{1} \cdots A_{k-1}B_{k}  + A_{1} \cdots A_{n}X_{0}
\end{equation}

From the condition that if the  expected value of the matrix $A_{1}$ is finite
\begin{equation}\label{eq4}
\mathbb{E} \log||A_{1}|| \leq \infty
\end{equation}
then  second term of Equation 3 converges to $0$ exponentially fast  $||A_{1} \cdots A_{n}|| \rightarrow 0$  and thus the distribution of the  word token generated at time $t$ converges to:

\begin{equation}\label{eq4}
R =  \Sigma_{k=1}^{\infty}A_{1} \cdots A_{k-1}B_{k}
\end{equation}

This shows the word token $X_{t}$ generated at the time $t$ is independent of the initial word $X_{0}$. This is of extreme importance, as violating this condition would make future word token highly sensitive to the initial word $X_{0}$. The distribution of the word generated is dependent on $R$. The renewal theory gives under reasonable conditions there exists $K > 0$ and $\eta > 0$ the probability of word $X_{t}$ follows \cite{cox1962renewal}:

\begin{equation}\label{eq4}
0 <  \lim_{x\to\infty} t^{\eta}P(|| X_{t}|| > t) < \infty
\end{equation}

Thus, our FPT model generates a word with long-range interactions because of the heavy-tail distributions. This heavy tail distribution introduces variance in the learning. So our model transcends from the memorization of the data-set to true learning during the fine-tuning process. This true learning or meta-learning capability is exploited for better generalizability during the fine-tuning process for binary classification.

\begin{figure}
    \centering
    \includegraphics[width=\columnwidth]{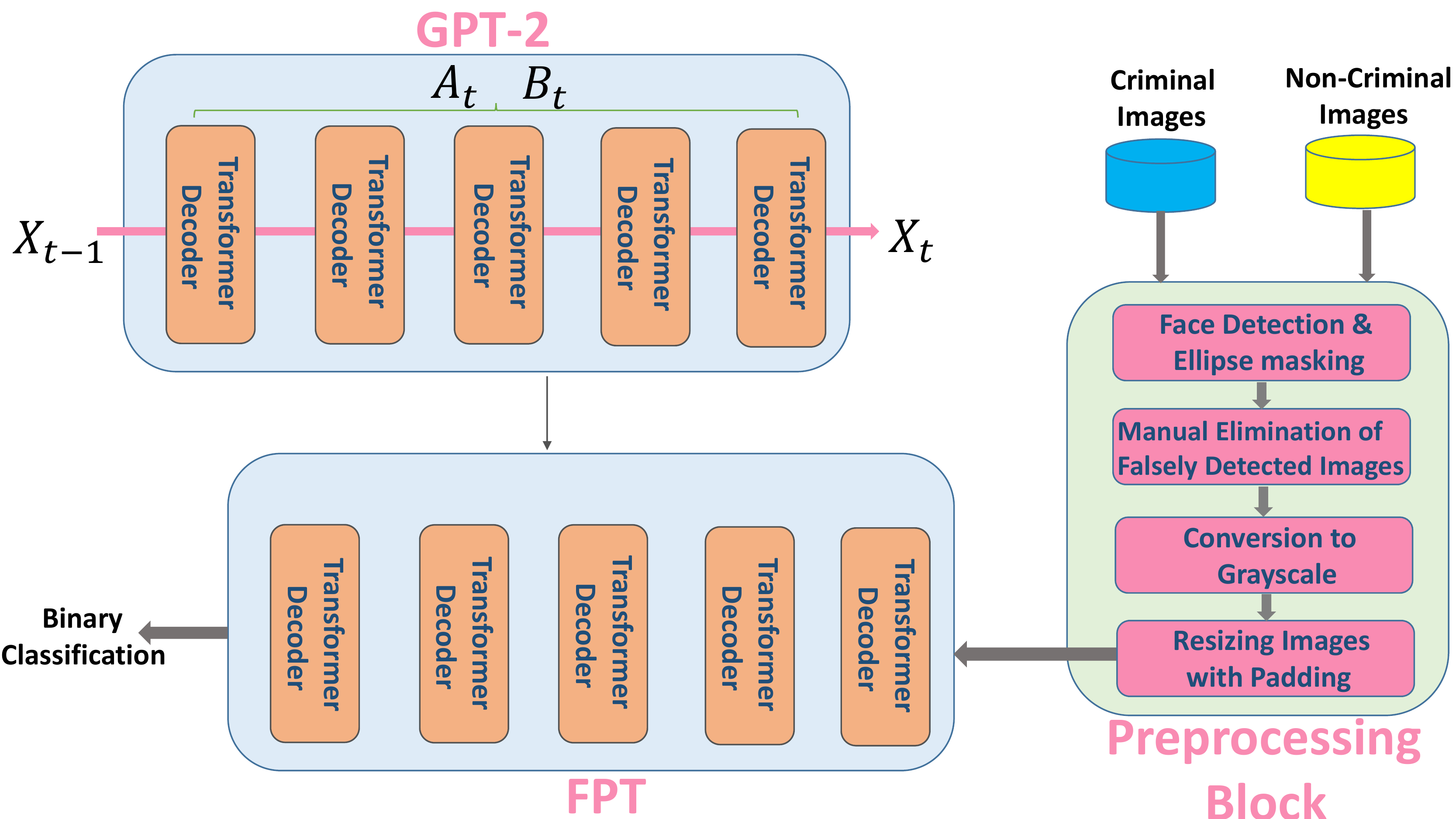}
    \caption{\textbf{Learning Pipeline for Binary Classification with FPT}}
    \label{fig:flowchart}
\end{figure}

\section{Dataset Description}\label{sec:datsetDescription}

We have collected facial image data from multiple resources through web crawling. The advantage of using multiple web sources to collect facial images is a balanced distribution between criminal and non-criminal classes. The facial dataset we have used from multiple web sources generates a mixed data distribution for training during the fine-tuning process. Thus, the bias embedded in the training set is reduced from multiple websites. We describe the method of data set collection in their prepossessing in detail (Section \ref{subsec:DatasetCollectPreprocess}). In the next section, we explain a Chaos-Based Image-Encryption algorithm \cite{al2012new} to encrypt facial images. The encrypted images are used as input images for training. The purpose of using encryption on facial images is to show that the pretrained transformers are powerful enough for binary classification. The practical usage comes from the fact that sharing criminal images across nations, continents etc are sensitive and have legal/ethical bindings, while sharing the encrypted images (without revealing the true identity) is more viable, as explained in Section \ref{subsec:DatasetEncryption}.

\subsection{Data Collection and Preprocessing}\label{subsec:DatasetCollectPreprocess}

\begin{table}
\centering
\begin{tabular}{|p{4.5cm}|l|l|} 
\hline
\textbf{Dataset} & \textbf{Total}   & \textbf{Pre-processed}\\\hline
\multicolumn{3}{|c|}{\textbf{\textcolor{blue}{Criminal Dataset}}}\\\hline
Smoking Gun           & 10,990 & 8,217\\\hline
National Institute of Standards and Technology (NIST)                  & 1,756  & 1,512\\\hline
Drug Enforcement Administration (DEA)                   & 587   & 337\\\hline
Crime stoppers        & 125   & 125\\\hline
Federal Bureau of Investigation (FBI)                   & 118   & 72\\\hline
Office of Inspector General (OIG)                   & 42    & 38\\\hline
U.S. Immigration and Customs Enforcement (ICE)                   & 41    & 12\\\hline
National Criminal Agency (NCA) & 22                      & 17\\\hline
Tennessee Bureau of Investigation (TBI)                   & 9     & 6\\\hline    
\multicolumn{3}{|c|}{\textbf{\textcolor{blue}{Non-Criminal Dataset}}}     \\ 
\hline
10k US Adult Faces Database & 10,168                      & 10,168\\\hline
Flickr-Faces-HQ Dataset (FFHQ) & 70,000                      & 168\\\hline
\end{tabular}
\caption{Criminal and Non-criminal dataset Collection Information.}
\label{table:CriminalDatasetInfo}
\end{table}

\begin{figure}
    \centering
    \includegraphics[width=0.9\columnwidth]{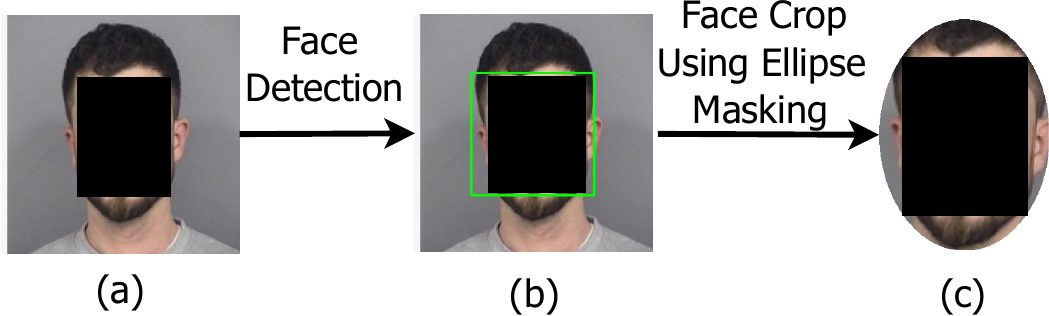}
    \caption{\textbf{Face Detection \& Image Crop Using Ellipse Mask: }In (a), we start by showing the original photo. The face is highlighted using the green box in (b) once detected. Finally, face is cropped using an ellipse mask (shown in (c)).}
    \label{fig:faceAlign}
\end{figure}

We have collected a total of 13,690 images of arrested or wanted criminals, and mugshots are collected from nine different sources (see Table \ref{table:CriminalDatasetInfo}, Column 1 and 2 for detail). From the collected images a total  of 11,934 RGB facial images are collected using web-scraping from eight sources (Smoking gun \cite{smokingGun}, Drug Enforcement Administration (DEA) \cite{dea}, Crime stoppers \cite{crimeStoppers}, Federal Bureau of Investigation (FBI) \cite{fbi}, Office of Inspector General (OIG) \cite{oig}, U.S. Immigration and Customs Enforcement (ICE) \cite{ice}, National Criminal Agency (NCA) \cite{nca}, Tennessee Bureau of Investigation (TBI) \cite{tbi}) and 1,756 gray-scale mugshot images of arrested individuals are obtained from National Institute of Standards and Technology (NIST) Special Database \cite{nist}. Images are in different formats PNG, JPG, or JPEG. The dataset contains images of individuals of various races, gender, and facial expression and contains both front and side (profile) views. Since we focus on frontal face shots, we need to eliminate profile views. Haar basis function-based cascade classifier detects images containing frontal face views and also detects the rectangular area containing the face\cite{viola2001rapid}. Images are first passed to a pre-trained version of this classifier, available in the OpenCV library in Python\cite{bradski2000opencv}. This is to select only the images containing frontal face views, and then we crop the rectangular area containing the face. Cropping the facial rectangle from the rest of the image prevents the classifier from being biased by peripheral or background effects surrounding the face. The non-frontal face images that are misclassified as frontal face images from the Haar feature-based cascade classifier are manually deleted. To further reduce the effect of peripheral or background surroundings, we crop the face using ellipse masking of the same size as shown in Figure \ref{fig:faceAlign}. The result contains approximately 10k front view face images of different formats (see Table \ref{table:CriminalDatasetInfo}, Column 3 for detail). We have also transformed image format to PNG and all RGB images to gray-scale to preserve consistency. To maintain the output dimension same as the input, we resize images with padding to 256×256.

A total of 10,168 RGB facial images are obtained from \textit{10k US Adult Faces Database} and are converted to the standard png format~\cite{bainbridge2013intrinsic}. We consider these images as noncriminal face shots. Images are sampled from different race, gender, and facial expressions. The final database  contains only front views and with reduced effect of peripheral and background surrounding through ellipse masking. We additionally added some images from Flickr-Faces-HQ Dataset (FFHQ) \cite{kazemi2014one} to balance both criminal and noncriminal datasets. FFHQ is a dataset consisting of human faces and includes more variation in terms of age, ethnicity, and image background. Since our focus is on frontal face shots, we need to eliminate profile views using the Haar feature-based cascade classifier\cite{viola2001rapid}. The non-frontal face images are then manually deleted. Also, to keep the age uniformity, images of the elderly and children are manually deleted from this dataset. 


\subsection{Data Encryption}\label{subsec:DatasetEncryption}
\begin{figure}
\begin{tabular}{cccc}
\subfloat[Criminal X encrypted image]{\includegraphics[width=0.22\columnwidth]{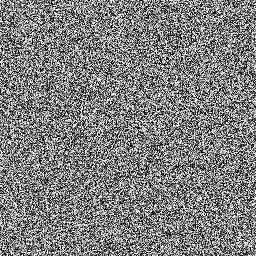}} &
\subfloat[Criminal X preprocessed image]{\includegraphics[width=0.22\columnwidth]{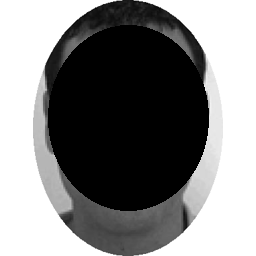}} &
\subfloat[Non-criminal Y encrypted image]{\includegraphics[width=0.22\columnwidth]{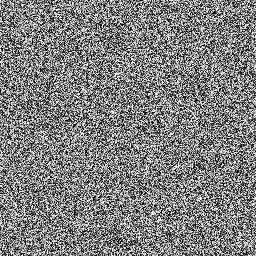}} &
\subfloat[Non-criminal Y preprocessed image]{\includegraphics[width=0.22\columnwidth]{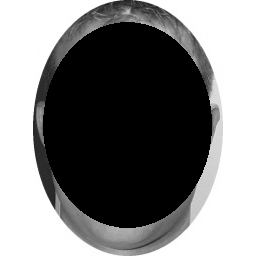}}
\end{tabular}
\caption{A few examples of encrypted and preprocessed criminal and non-criminal images.}
\label{fig:EncryptedImagesExample}
\end{figure}

We have used Chaos-Based Image-Encryption algorithm to encrypt the  images\cite{al2012new}. Chaos-based encryption uses some dynamical systems method to generate random sequences of numbers. This sequence is then used to generate public key for the encryption. Chaotic systems are sensitive on initial conditions, similarity to random behavior. For the key generation a Chaotic map $T_{p}$ of degree $p$ is defined  by a recurrent relation (e.g. Chebyshev map)~\cite{kocarev2011chaos}:
 \begin{equation}\label{eq3}
T_{p+1}(x) = 2xT_{p}(x) -T_{p-1}(x)
\end{equation}
In order to generate keys we first generate a large integer $s$ and a random number $x\in [-1, 1]$ and then compute the chaotic map $T_{s}(x)$. The public key is the pair $(x, T_{s}(x))$ and private key is $s$. Now in order to  encrypt facial image $M$ we again generate a large integer $r$ and compute the  following functions $T_{r}(x), T_{r.s} = T_{r}(T_{s}(x)),  X = M.T_{r.s}(x)$. The encrypted image is then $C = (T_{r}(x), X)$ on which our learning  FPT does the binary classification.  We view our chaos-based encryption on the facial image as a non-linear transformation on the pixel space. We have shown some examples of chaos-based encrypted images and their corresponding pre-processed images in Figure~\ref{fig:EncryptedImagesExample}. Though these images are indistinguishable and unidentifiable for human eyes, our pretrained transformer FPT exploits this equivariance on encrypted images to classify~\cite{olah2020naturally}. The high accuracy of binary classification on the encrypted space points us to an exciting direction of the equivariance principle deep models use. The practical advantage of working in the encrypted pixel space comes from sharing sensitive (e.g. criminal images) are not always possible. Such sharing of data has both legal and ethical concerns. Sharing encrypted images across the boundary will alleviate the ethical and legal issues. The transnational police agencies e.g. Interpol Europol that works accross the nations will be able to run smoothly because of the encrypted images.

\section{Architecture}\label{sec:Architecture}

The model we use in our experiments is a frozen pretrained transformer (FPT) model of GPT-2 model\cite{lu2021pretrained}. Our FPT model belongs to the Transformer class of architecture introduced by Ashish Vaswani et al~\cite{vaswani2017attention}. Our choice of FPT, which is a decoder only block transformer because of its generative capability of single word at a time. This recursive generation of words, along with the forward masking gives the model heavy-tail properties. This heavy-tail gives the ability to understand the pixel content. The architecture of FPT has the following attributes~\cite{radford2019language,wolf2019huggingface}: embedding size is $768$, the number of layers is $12$, output dimension is  $2$. FPT is pretrained (for word token generation) on a large corpus of $\sim$40 GB of text data. The pretrained model is then fine-tuned with 20K frontal images as explained in the ~\ref{subsec:DatasetCollectPreprocess}. During the FPT finetuning only the linear input, output layer,  positional embeddings, and layer norm parameters are updated during the stochastic gradient descent method. The rest of the architecture is not updated and thus remains frozen.

\section{Results}\label{sec:Results}
In this section, we first describe the performance metrics to measure the generalization capacity of the FPT (in Section \ref{subsec:performaceMeasures}), and then we discuss the results of binary classification (in Section \ref{subsec:results}). We finetune our model with approximately 20k images (criminals plus non-criminals) in the original image and its encrypted form separately as shown in the Table~\ref{tab:TrainTestData}. A total of 16,672 trained images and 4,000 test images are used in our experiments, the results of which are reported in Section \ref{subsec:results}. Our experiments are conducted on NVIDIA Tesla V100 GPU and 512GB memory.

\begin{table}
\centering
\begin{tabular}{|c|c|c|} 
\hline
              & Criminal & Non-Criminal  \\ 
\hline
Train dataset & 8336     & 8336          \\ 
\hline
Test dataset  & 2000     & 2000          \\
\hline
\end{tabular}
\caption{Train-test data information.}
\label{tab:TrainTestData}
\end{table}

\subsection{Measuring Performance}\label{subsec:performaceMeasures}
\subsubsection{AUC (AUROC)}
The area under the receiver operating characteristic (AUROC) is a performance metric used to evaluate classification models. The AUROC is the probability that a randomly selected positive example has a higher predicted probability of being positive than a randomly selected negative example. The AUROC is calculated as the area underneath a curve that measures the trade-off between true positive rate (TPR) and false positive rate (FPR) at different decision thresholds d. For balanced two-class data, a random baseline classifier has an AUROC as 0.5 (d = 0.5), while a perfect classifier has an AUROC of 1.0.

\subsubsection{Average Precision}
Average precision or AUPRC (Area Under the Precision-Recall Curve) is calculated as the area under Precision-Recall (PR) curve. A PR curve shows the trade-off between precision and recall across different decision thresholds. Thus, the average precision is high when a model can correctly handle positive examples. With AUPRC, the baseline is equal to the fraction of positive examples. The fraction of positive examples is calculated as the number of positive examples divided by the total number of examples. This gives different classes different AUPRC baselines. For balanced two-class data, the AUPRC baseline is 0.5, while a perfect classifier has an average precision of 1.0.

\begin{figure}
  \centering
  \subfloat[Pre-processed data: Accuracy]{\label{figur:preprocessedAcc}\includegraphics[width=0.4\columnwidth]{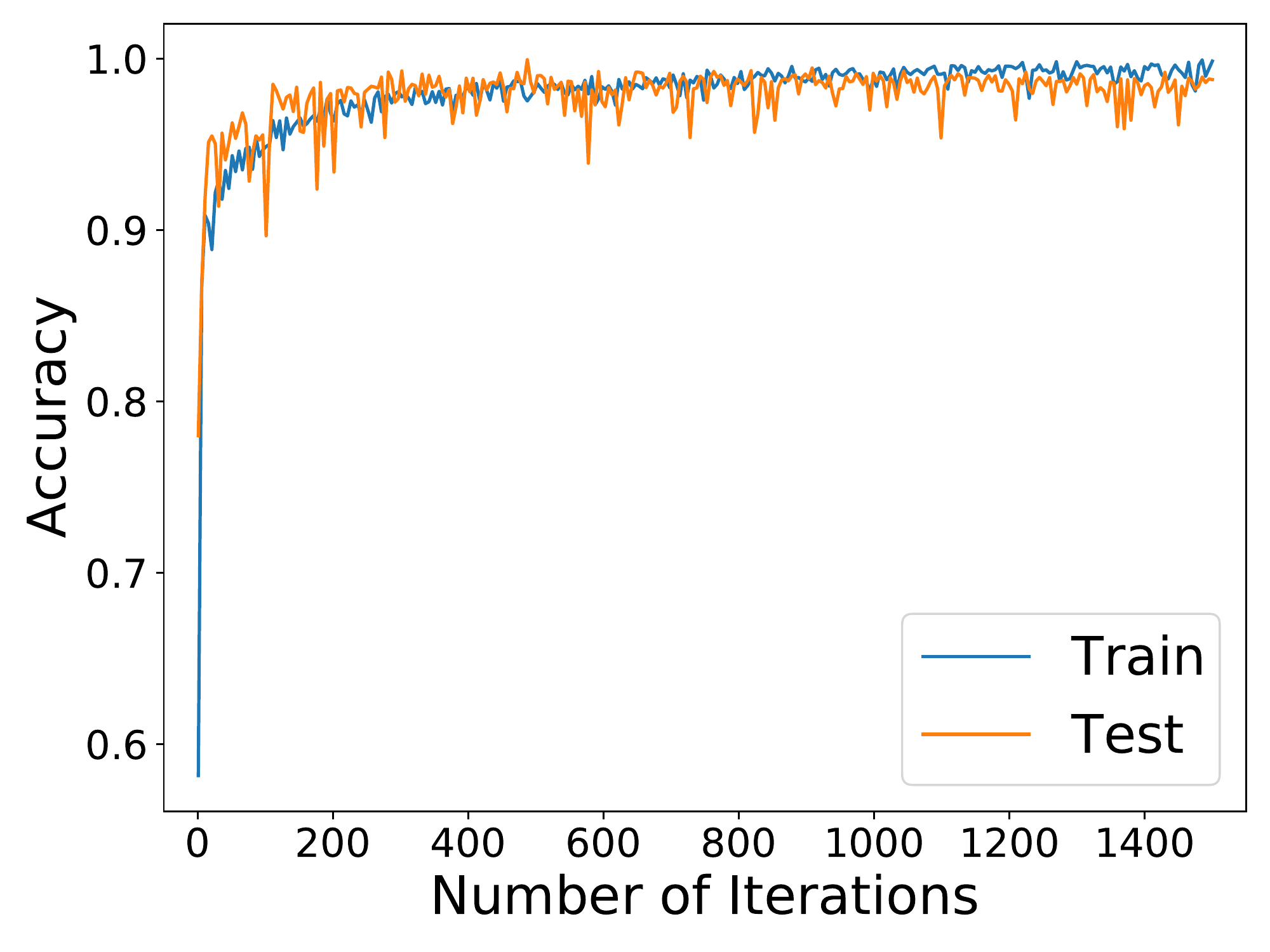}}
  \hspace{3mm}
  \subfloat[Pre-processed data: Loss]{\label{figur:preprocessedLoss}\includegraphics[width=0.4\columnwidth]{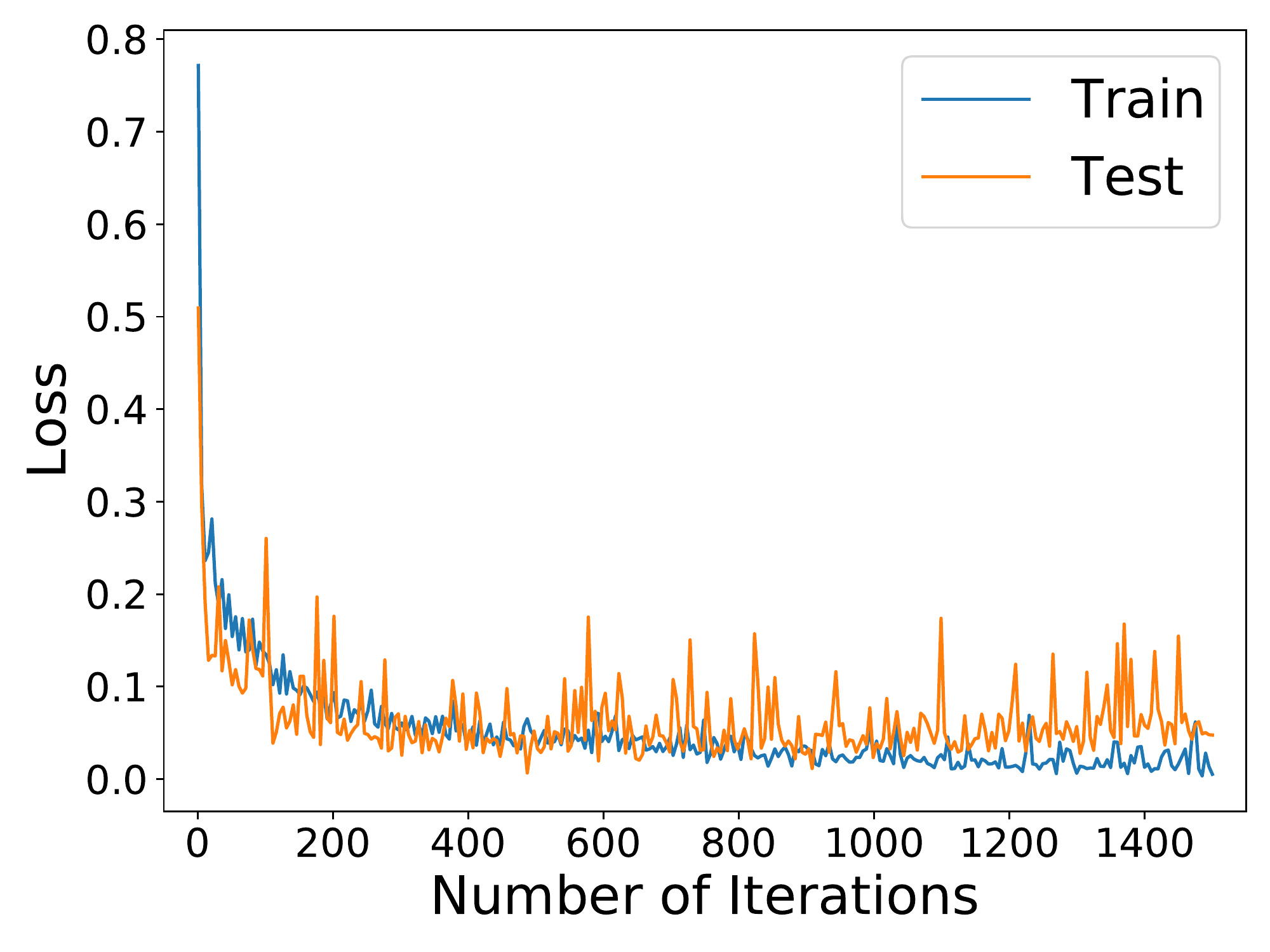}}\\
    \subfloat[Encrypted data: Accuracy]{\label{figur:EncryptAcc}\includegraphics[width=0.4\columnwidth]{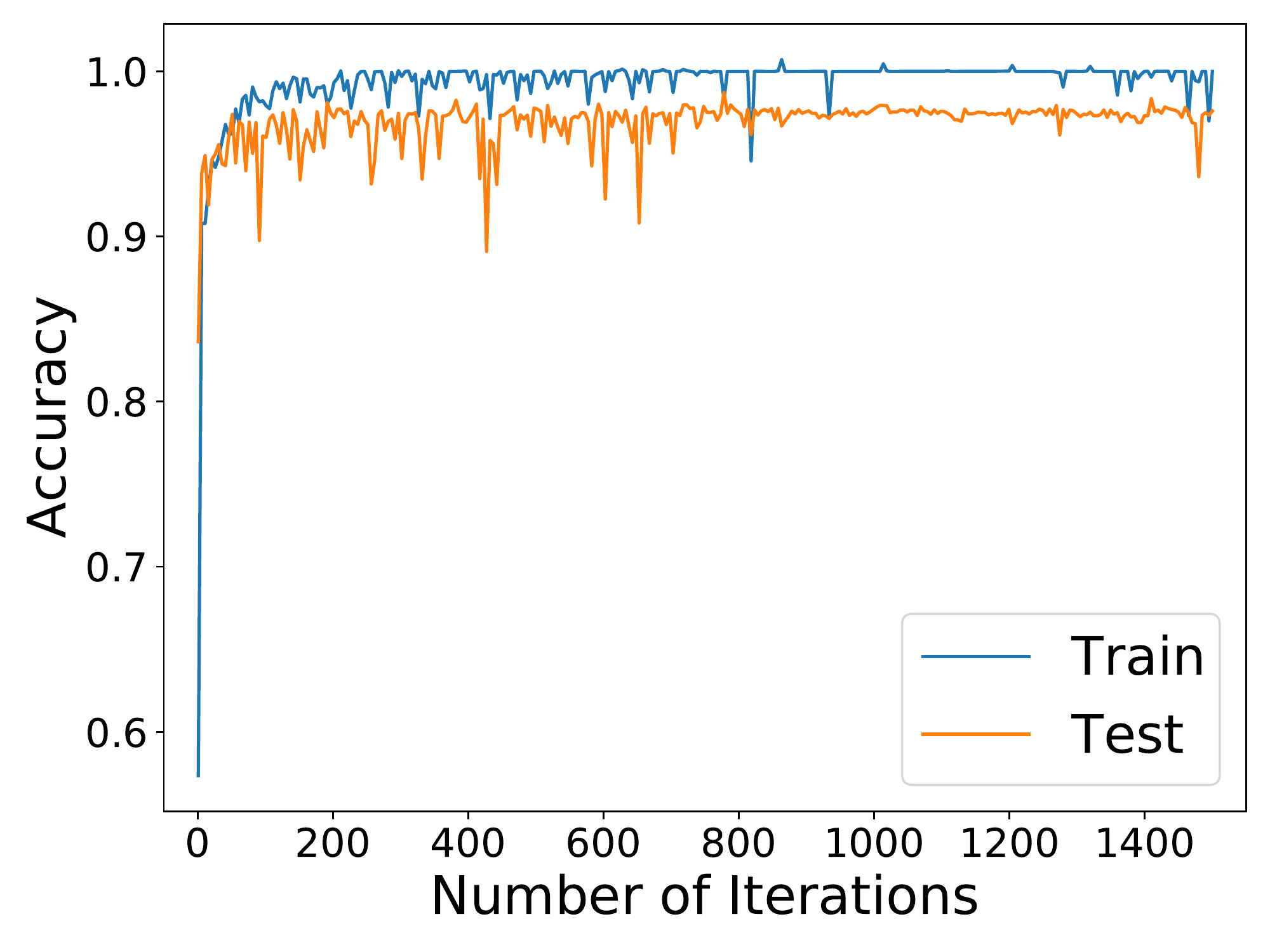}}\hspace{3mm}
  \subfloat[Encrypted data: Loss]{\label{figur:EncryptLoss}\includegraphics[width=0.4\columnwidth]{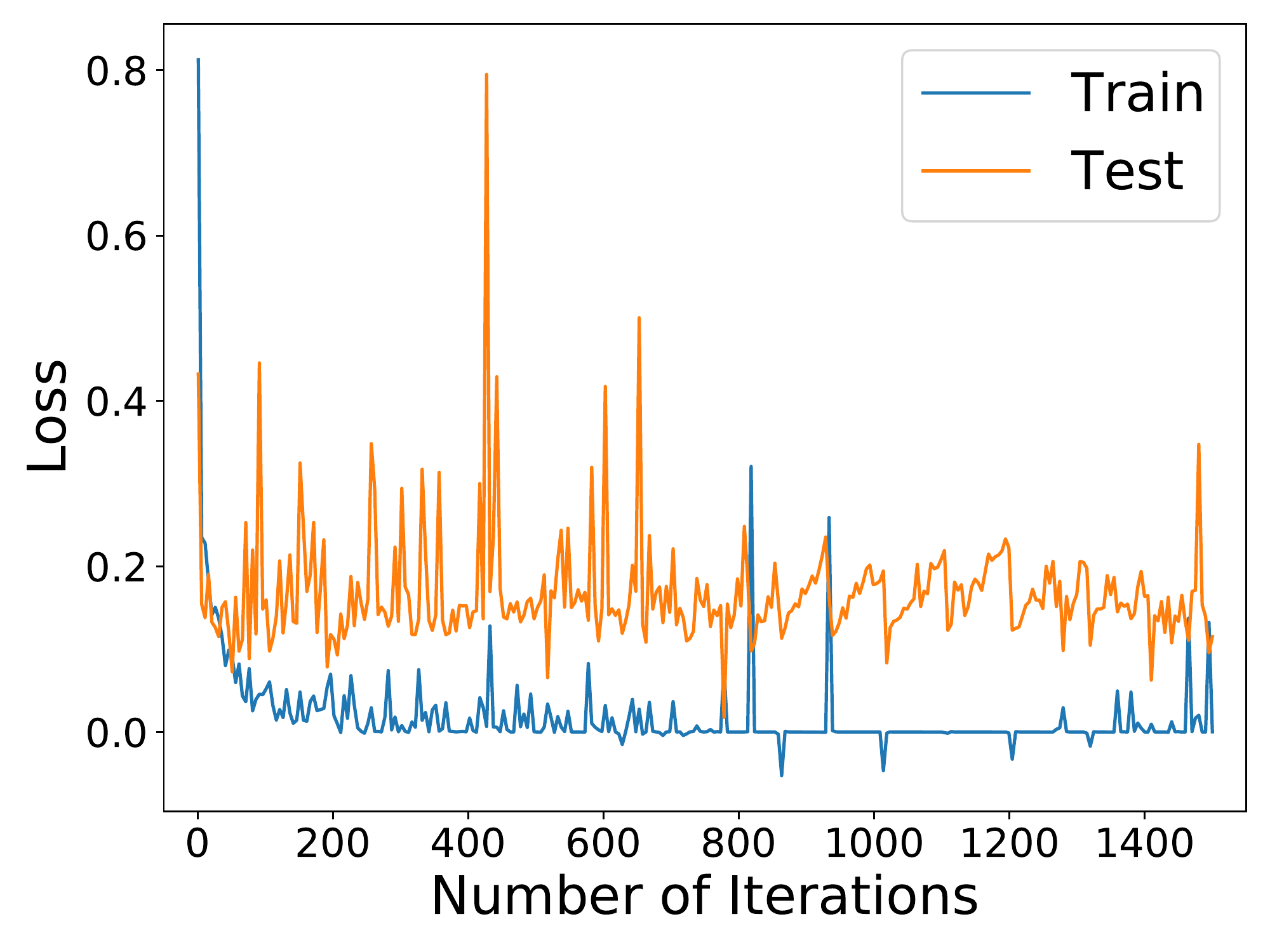}}
  \caption{Train-test accuracy and loss for pre-processed facial images (a, b) and encrypted facial images (c, d) respectively.}
\label{fig:acc_loss_complete_data}
\end{figure}

\begin{figure}
  \centering
  \subfloat[Pre-processed data]{\label{fig:AUROC_AUPRC_1}\includegraphics[width=0.6\columnwidth]{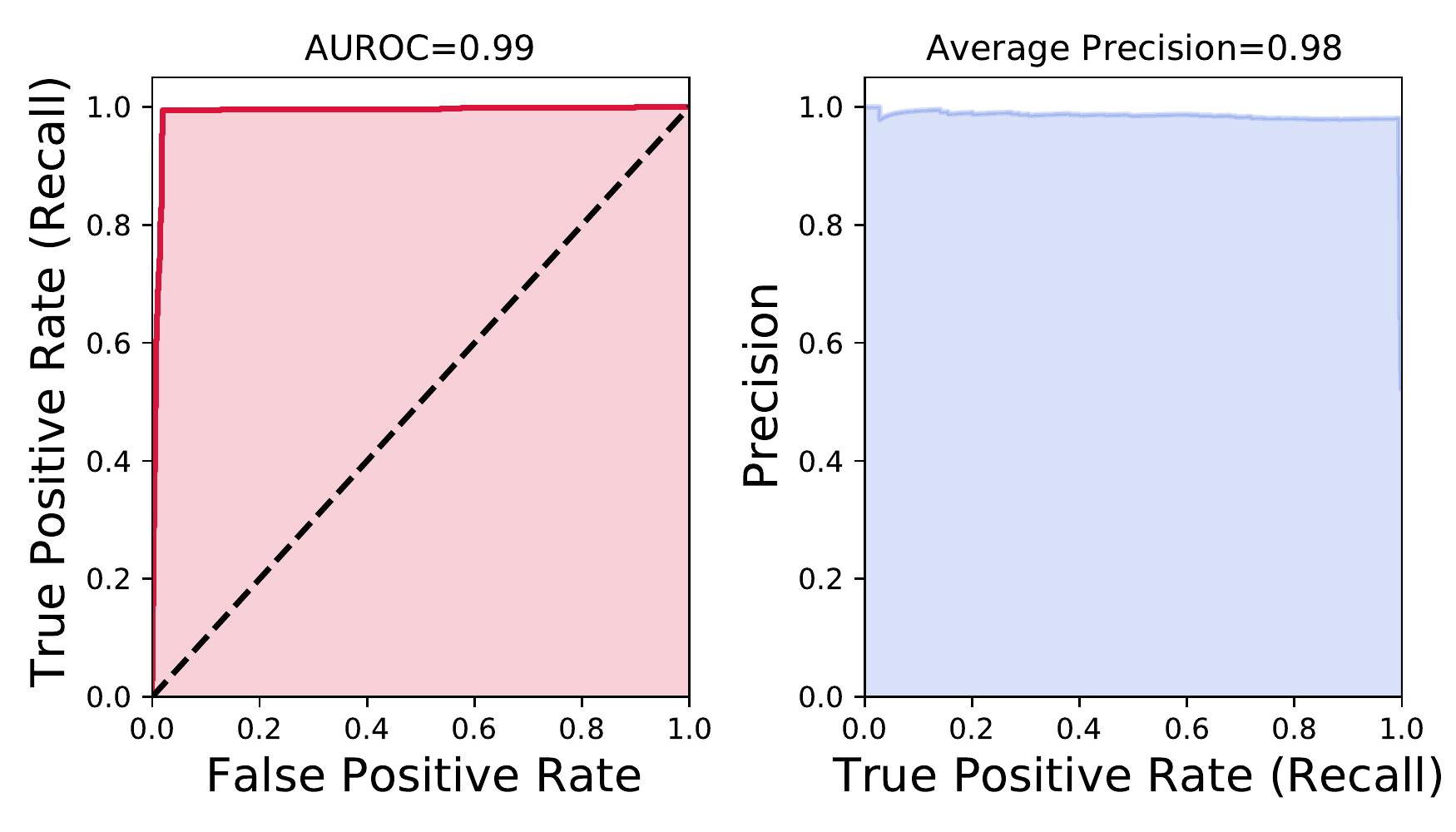}}\\
  \subfloat[Encrypted data]{\label{fig:AUROC_AUPRC_2}\includegraphics[width=0.6\columnwidth]{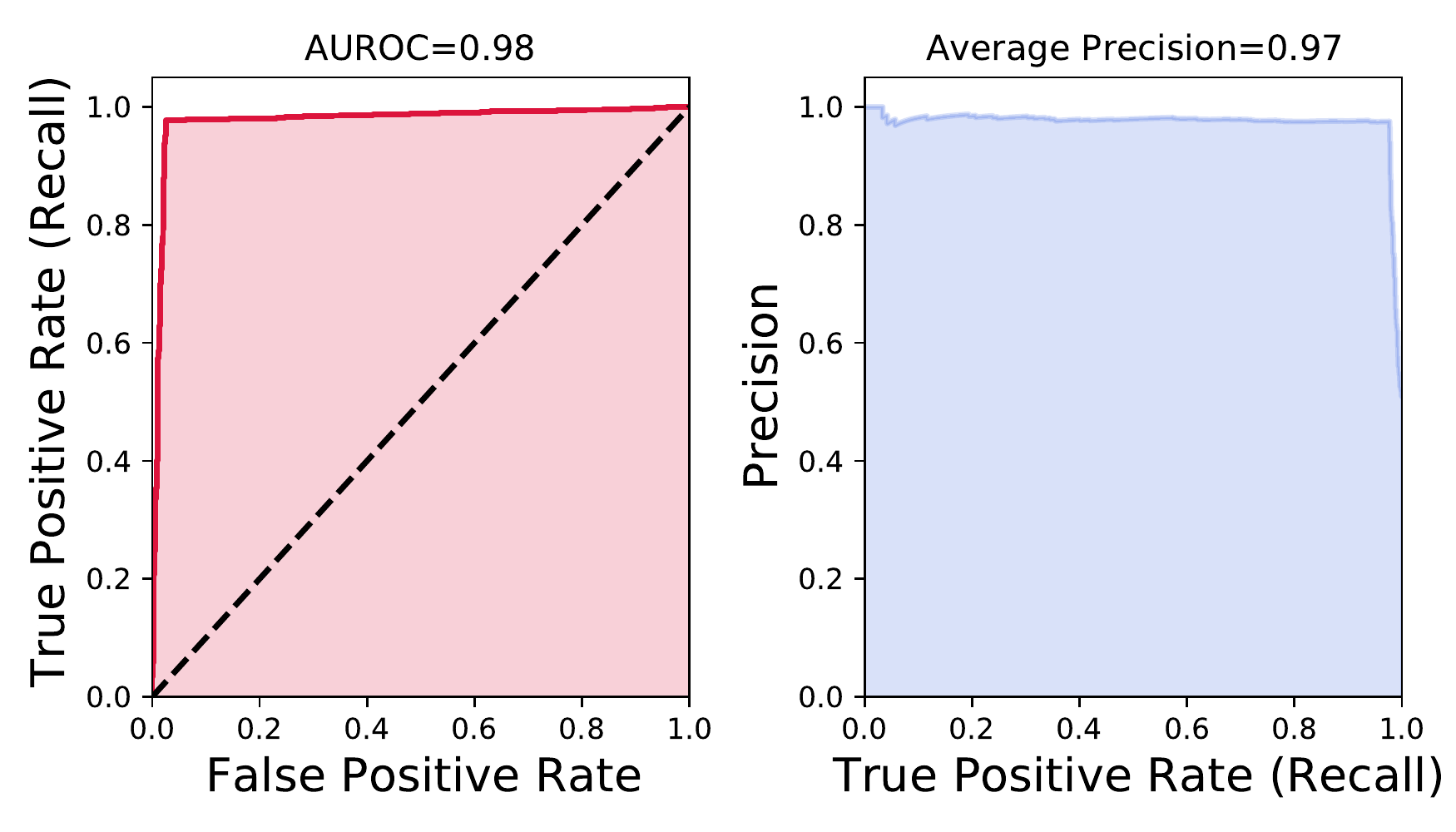}}
  \caption{AUROC and AUPRC plots for pre-processed, and encrypted data.}
\label{fig:AUROC_AUPRC}
\end{figure}

\subsection{Generalization Results}\label{subsec:results}
We have used two sets of datasets: (1) pre-processed frontal facial images, (2) chaos-based encrypted images, and fine-tuned two separate FPT models. Our models' train and test accuracies and  losses with epochs are plotted in Figure \ref{fig:acc_loss_complete_data}. Figure \ref{figur:preprocessedAcc}, and \ref{figur:preprocessedLoss} shows the accuracy and loss of fine-tuned FPT model on preprocessed frontal image data.  Figure \ref{figur:EncryptAcc}, and \ref{figur:EncryptLoss} shows the accuracy and loss of fine-tuned FPT model on encrypted data. For pre-processed data, we can infer that the model is trained properly as the test and train accuracies reach to maximum in around 350-400 epochs and corresponding loss also saturates to the minimum value in similar number of epochs (see Figure \ref{figur:preprocessedAcc}, and \ref{figur:preprocessedLoss}). On the other hand, for the encrypted data, the training accuracy is always higher than the test accuracy even after 1500 epochs, and corresponding loss is always higher for test data. Even though we can explain the case of encrypted images as a standard case of overfitting from the statistical learning theory principles, but from the \textit{scaling hypothesis}. The billion parameter FPT model is trained in an unsupervised manner on a large internet-scale text for word generation. While trained to generate text, our FPT has no choice but to solve many hard problems. This drives our FPT to go for meta-learning. Our FPT is forced from memorizing parts of the data during pretraining to the true learning. Thus, our FPT builds informative priors while pretraining on the large text data set. The FPT  as a meta-learner learns as a amortized Bayesian inference model. Even though in our paper we show the accuracy for binary classification, we propose two hypotheses based on our observation: 

\begin{itemize}
  \item Accuracy won't suffer much even for the problem of  multi-label classification of the facial images with FPT.  
  \item We will further improve the  training and accuracy of the encrypted images, by choosing carefully the initial conditions of Chaotic map (Chebyshev map) in chaos-based encryption.
\end{itemize}

\begin{figure}
  \centering
  \subfloat[Pre-processed data]{\label{figur:11}\includegraphics[width=0.5\columnwidth]{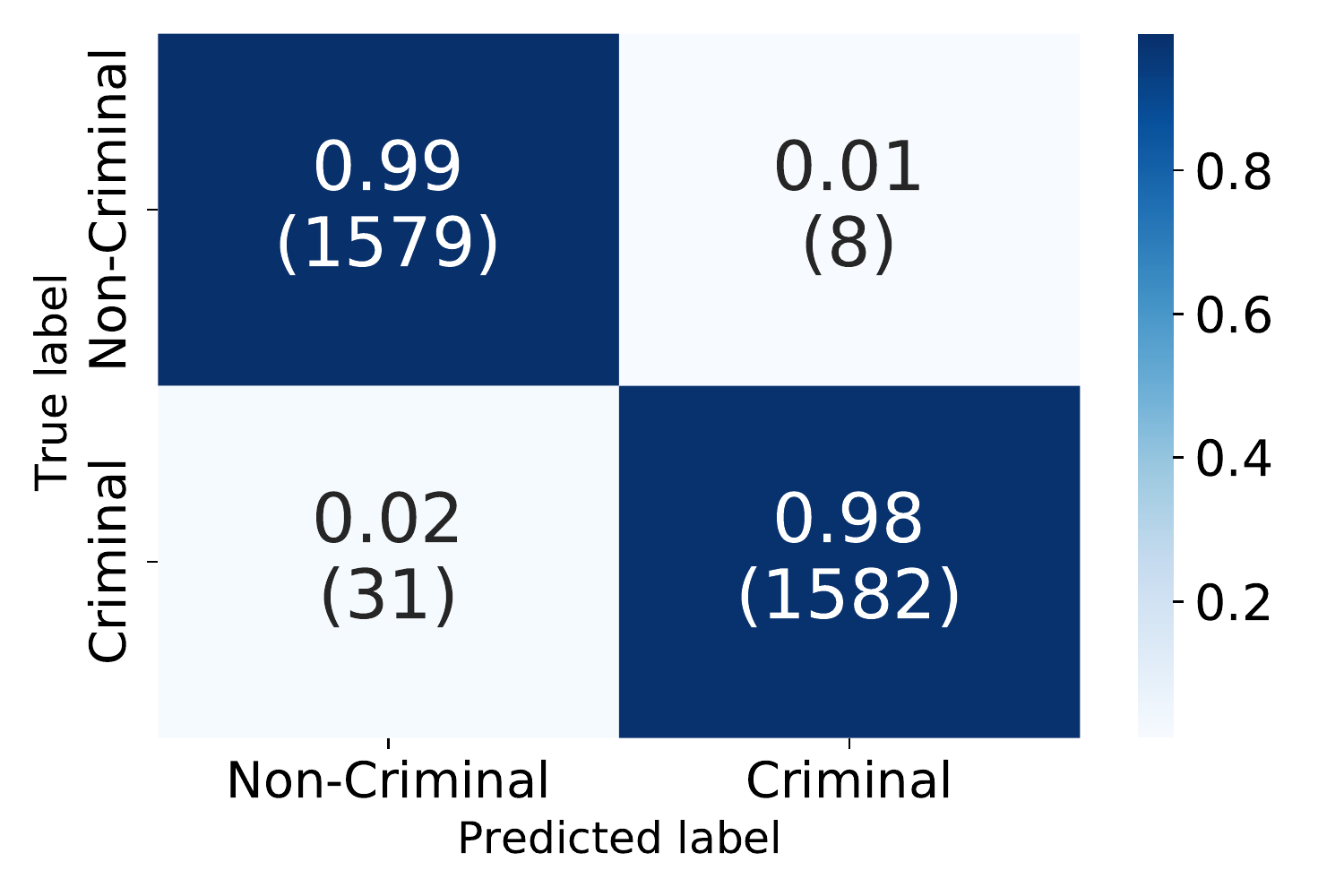}}
  \subfloat[Encrypted data]{\label{figur:12}\includegraphics[width=0.5\columnwidth]{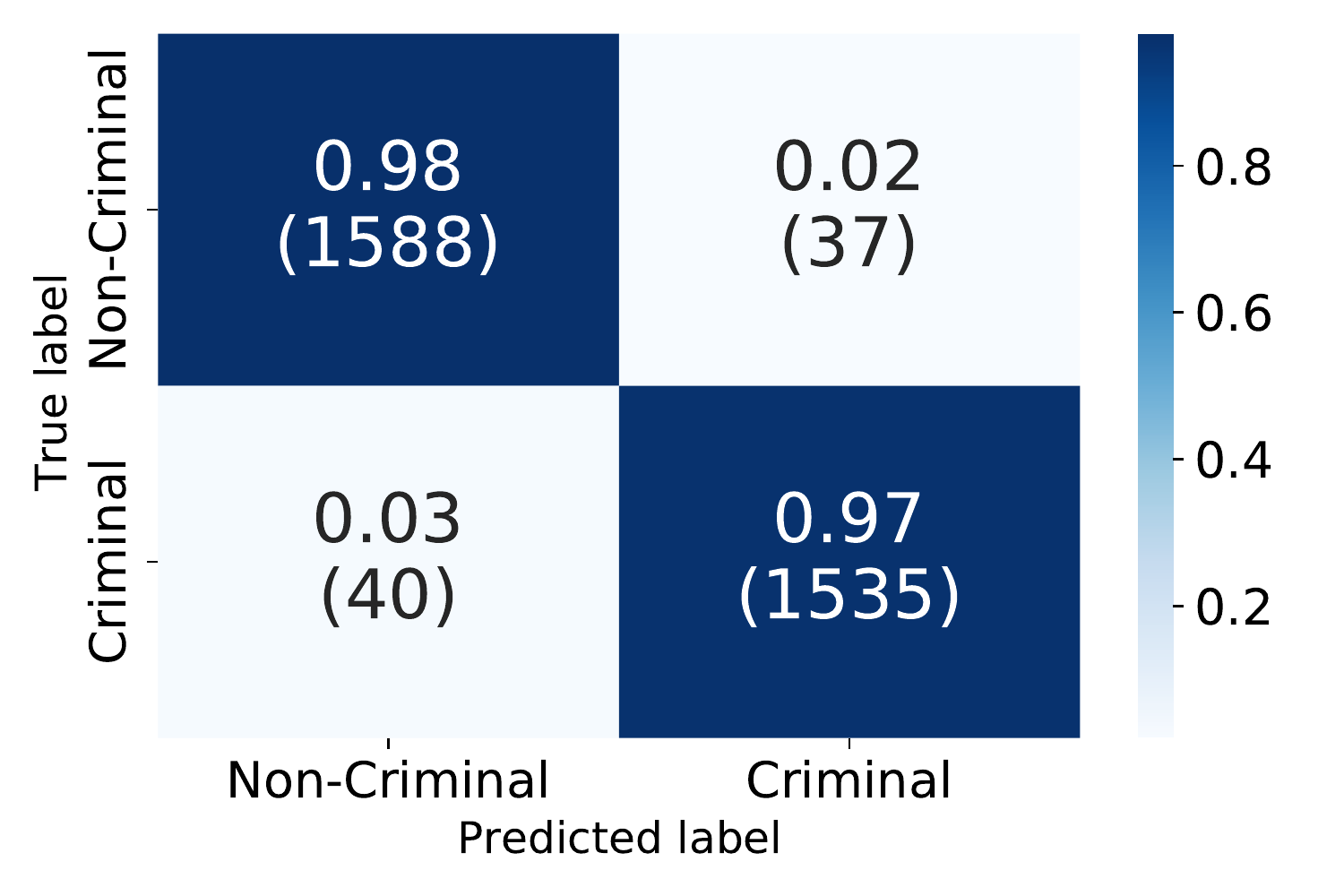}}
  \caption{Normalized confusion matrix}
\label{fig:normalized-confusion-matrix-1}
\end{figure}

Next, we calculate the performance of our fined tuned FPT models for pre-processed facial images and encrypted images separately using two metrics, AUROC, and Average precision. FPT generates many true positives and true negatives during the testing as our AUROC, and average precision are both high (Figure \ref{fig:AUROC_AUPRC}). In the ROC plot (red), we observe the decision thresholds d = 0.9 to d = 0.5 span a small interval of $\text{FPR} = f_{p}/(f_{p}+t_{n})$. Because of high decision thresholds, the $f_{p}$ are low, and the $t_{n}$ are high, thus producing a small FPR. The average precision is not improved by the number of true negatives, as true negatives are not used in the calculation. The average precision is improved by the decrease in false positives. This is because some examples are shifted from false positives to true negatives. This is in order to keep the dataset balanced. Precision = tp/(tp+fp), so when we make the fp smaller, we increase the precision. We use a confusion matrix as shown in Figure \ref{fig:normalized-confusion-matrix-1} that calculates how many images from both criminal and non-criminal classes got correctly classified. We can observe from here that 0.02\% of images from criminal and 0.01\% of from non-criminal classes are getting misclassified using preprocessed images. Similarly, for the encrypted data, we can observe that 0.03\% of images from criminal and 0.02\% of images from non-criminal classes are getting misclassified. 


\section{Conclusion}
We have presented a novel method of using a frozen pretrained transformer (FPT) trained on the text corpus and then fine-tuned on the images for binary classification. We have shown that large language models of a transformer are expressive enough in solving vision tasks. We have used a pretrained transformer model trained to generate text and use this pretrained transformer to solve the binary classification of facial images. The high accuracy of our pretrained model on the classification task shows the phenomena of "\textit{blessings of scale}. Our FPT with billions of parameters while doing the pretraining of word generation started showing its meta-learning capability. Because of this phenomena's blessings of scale, our FPT becomes powerful and more generalizable while solving the image classification problem\cite{huang2021textrmimplicit}. The other surprising capacity of our model is its ability to classify even on the encrypted images. Our work contributes towards using 
transformer models in the privacy-preserving machine learning method. 

On the practical side, our paper contributes a step towards an automated classification of criminal images with high accuracy. These models can be deployed by police and defense agencies, border security for airports, railway stations, and other important and crowded venues as a pre-crime tech tool. The other important side we address is the inherent bias by the people in the security in convicting people wrongly. Social movement like \textit{Black Lives Matter} highlights the police brutality towards particular sections in the society. We need to build more ethical tools that are free from human bias, and one crucial way is to build large machine learning models and trained on a wide variety of data. This helps the model not to have a myopic view about the world and judge humans ethically.

\section*{Ethical Statement}
 
While collecting and analyzing the data set for this investigation, the researchers were committed to high ethical standards. All researchers were guided by three ethical principles throughout the various stage of the research, including data collection, data processing, data analysis, and dissemination of the key findings:
\begin{enumerate}
\item The researchers ensured the quality and integrity of all research practices.
\item Personal information contained in the collected data has been treated as confidential and was anonymized in public presentations and publications. 
\item The researchers testify that their research is independent and impartial. 
\end{enumerate}

Throughout the project, the researchers considered both positive and negative outcomes of the research. For instance, the publication of outcomes, such as the identification of people in the dataset, could possibly affect their well-being. To mitigate such risks, the researchers anonymized any personal information in published materials to elicit informed consent on the draft paper. Finally, the data sets used for the investigation were stored at a protected locations in the cloud, and all personal information gathered throughout the research will be destroyed after the last publication of the project.

\section*{Acknowledgement}
This research (or publication) has been financed by the Europan Social Fund, Estonian Research Council Institutional Research grant PUT PRG306 and ERDF via the IT Academy Research Programme, and H2020 framework project, SoBigData++ and CHIST-ERA project SAI. This work has also been supported by NASA (Awards 80NSSC20K1720 and 521418-SC), NSF (Awards 2007202 and 2107463).

\bibliographystyle{plain}
\bibliography{sample-base}  
%
%


\end{document}